\documentclass[11pt,a4paper]{article}
\usepackage{times,latexsym}
\usepackage{url}
\usepackage[T1]{fontenc}

\usepackage[]{emnlp2020}

\usepackage{times}
\usepackage{latexsym}
\usepackage{booktabs}
\usepackage{color}
\usepackage{graphicx}
\usepackage[TABBOTCAP]{subfigure}
\usepackage[shortlabels]{enumitem}
\usepackage{algorithm}
\usepackage{algpseudocode}

\usepackage{url}

\newcommand{\ignore}[1]{}

\newcommand{\APSI}[0]{\texttt{APSI}}

\usepackage{multirow}

\newcommand{\muhao}[1]{\textcolor{blue}{MC:{#1}}}

\def\CM{{\mathcal C}}

\def\EM{{\mathcal E}}

\def\GM{{\mathcal G}}

\def\C{{\bf C}}

\def\E{{\bf E}}

\def\0{{\bf 0}}
\def\1{{\bf 1}}

\usepackage{xspace,mfirstuc,tabulary}

\def\aclpaperid{1421} 
\aclfinalcopy 
\newif\iftaclinstructions
\taclinstructionsfalse 
\iftaclinstructions
\renewcommand{\confidential}{}

\newcommand{\instr}
\fi 

\title{Analogous Process Structure Induction for Sub-event Sequence Prediction}
\author{Hongming Zhang$^1$\thanks{$\;$ This work was done when the first author was visiting the University of Pennsylvania.}$\;$, Muhao Chen$^2$, Haoyu Wang$^2$, Yangqiu Song$^1$, \& Dan Roth$^2$\\
$^1$Department of Computer Science and Engineering, HKUST\\
$^2$Department of Computer and Information Science, UPenn\\
\texttt{\{hzhangal, yqsong\}@cse.ust.hk}\\
\texttt{\{muhao, why16gzl, danroth\}@seas.upenn.edu}\\  \\ 
}

\date{}

\begin{document}
\maketitle
\begin{abstract}
Computational and cognitive studies of event understanding suggest that identifying, comprehending, and predicting events depend on having structured representations of a sequence of events and on conceptualizing (abstracting) its components into (soft) event categories. Thus, knowledge about a known process such as ``{\em buying a car}'' can be used in the context of a new but analogous process such as ``{\em buying a house}''. Nevertheless, most event understanding work in NLP is still at the ground level and does not consider abstraction. In this paper, we propose an Analogous Process Structure Induction (\APSI) framework, which leverages analogies among processes and conceptualization of sub-event instances to predict the whole sub-event sequence of previously unseen open-domain processes. As our experiments and analysis indicate, \APSI\footnote{Code is available at: \url{http://cogcomp.org/page/publication_view/910}.}~supports the generation of meaningful sub-event sequences for unseen processes and can help predict missing events.

\ignore{
Humans are capable of understanding daily life processes by their sub-event sequences, and such knowledge can help people finish many cognitive tasks such as event prediction ~\cite{zacks2007event} and transfer their knowledge about a known process (e.g., `buy a car') to a new but analogous process (e.g., `buy a house'). 
In this paper, 
we propose an Analogous Process Structure Induction (APSI) framework, which relies on the analogous property of processes as well as the conceptualization and instantiation of events, to prediction the whole sub-event sequences for new unseen open-domain processes.
Intrinsic experiments and detailed analysis demonstrate that even though the proposed task is challenging, our framework can outperform all baseline methods and generate meaningful sub-event sequences for unseen processes.
Moreover, extrinsic experiments also show that the induced process knowledge can help predict missing events.
}

\end{abstract}

\section{Introduction}


Understanding events has long been a challenging task in NLP, to which many efforts have been devoted by the community.
However, most existing works are focusing on 
procedural (or \emph{horizontal}) event prediction tasks. Examples include predicting the next event given an observed event sequence~\cite{DBLP:conf/www/RadinskyDM12} and identifying the effect of a biological process (i.e., a sequence of events) on involved entities~\cite{DBLP:conf/emnlp/BerantSCLHHCM14}.
These tasks mostly focus on predicting related events in a procedure based on their statistical correlations in previously observed text.
As a result, understanding the meaning 
of an event might not be crucial for these \emph{horizontal} tasks.
For example, simply selecting the most frequently co-occurring event can 
offer acceptable performance on the event prediction task~\cite{DBLP:conf/aaai/Granroth-Wilding16}.

\begin{figure}
    \centering
    \includegraphics[width=0.9\linewidth]{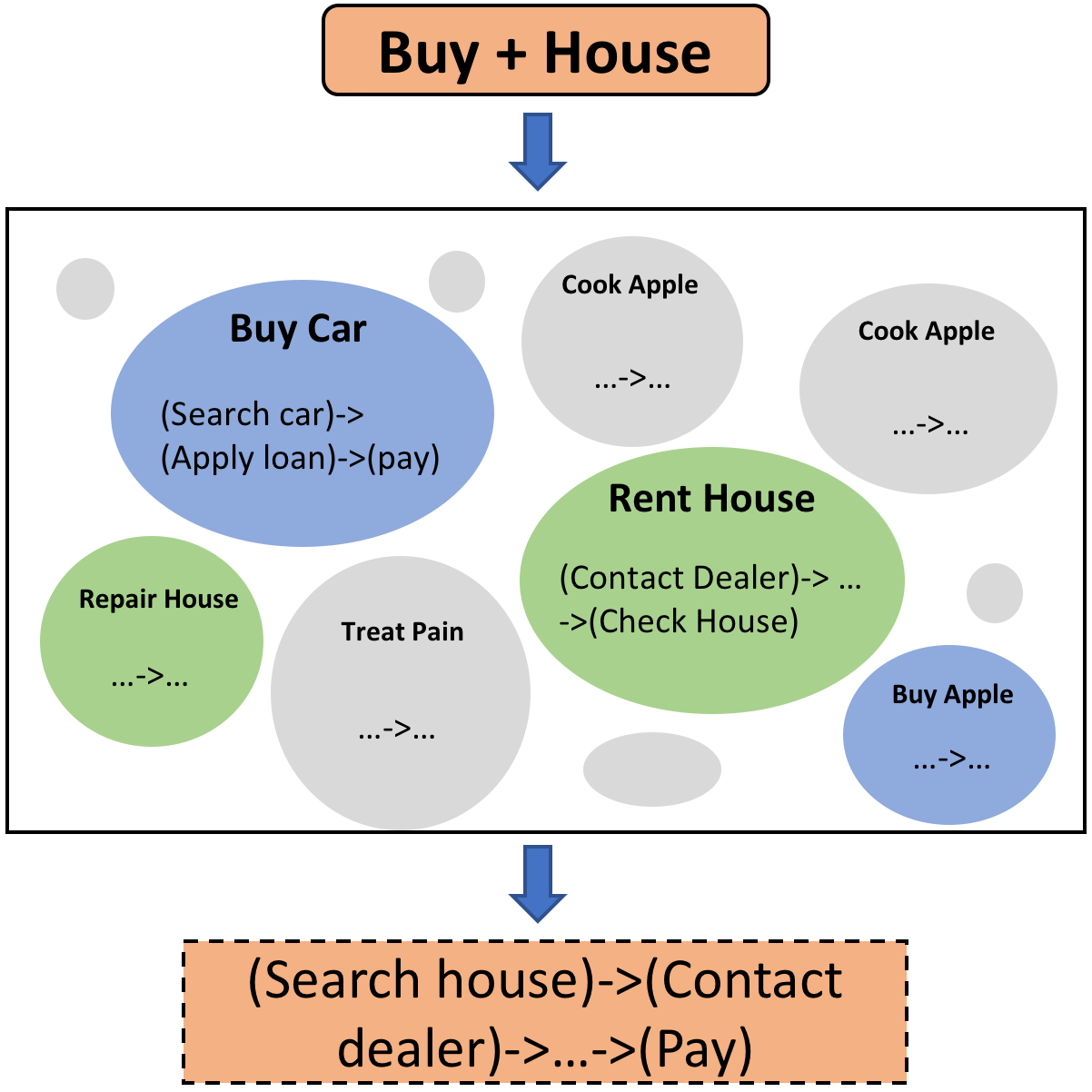}
    \caption{An illustration of leveraging known processes to predict the sub-event sequence of a new process.
    }
    \label{fig:intro_example}
\end{figure}

\begin{figure*}
    \centering
    \includegraphics[width=\linewidth]{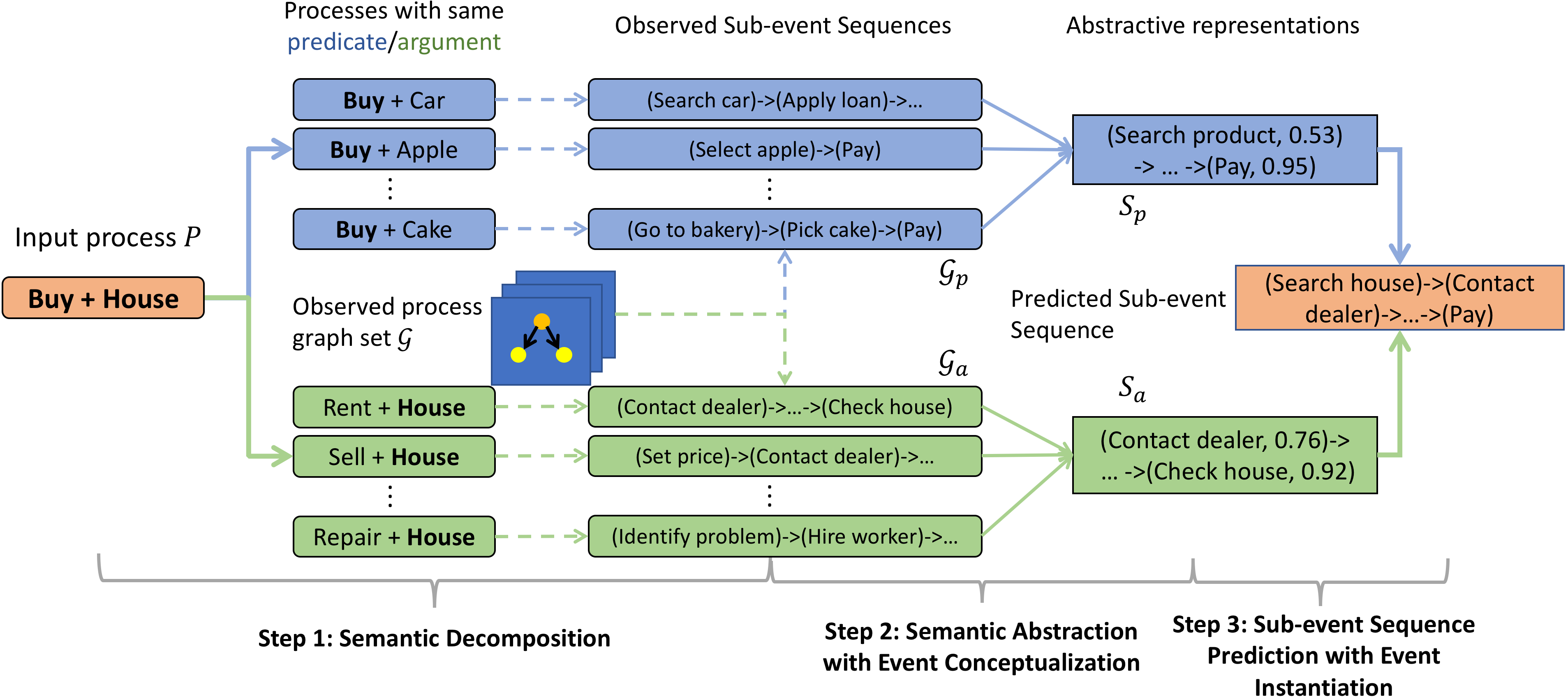}
    \caption{
    Demonstration of the proposed APSI framework. Given a target process $P$, we first decompose its semantics into two dimensions (i.e., predicate and argument) by grouping 
    processes that share
    a 
    predicate or an argument. 
    For each such group of processes, we then leverage the observed process graphs $\GM$ to generate an abstract and probabilistic representation for their sub-event sequences.
    In the last step, we merge them with an instantiation module to produce the sub-event sequence of $P$.}
    \label{fig:framework}
\end{figure*}

Computational and cognitive studies~\cite{schank1977scripts,zacks2001event} suggest that inducing and utilizing the
hierarchical structure\footnote{The original paper refers to the knowledge about processes and their sub-events as event schemata.} of 
events is a crucial component of how humans understand new events and can help many aforementioned \emph{horizontal} event 
prediction tasks.
Consider the example in Figure~\ref{fig:intro_example}.
Assume that one has never bought a house, but is familiar with how to ``buy a car'' and ``rent a house''; 
referring to analogous steps in these two relevant processes would still provide guidance for the target process of ``buy a house''. 
Motivated by this hypothesis, 
our work proposes to directly evaluate a model's event understanding ability. 
We define this as the ability to identify {\em vertical} relations, that is, to predict
the sub-event sequence of a new process\footnote{A process is a more coarse-grained event by itself. We use this term to distinguish it from sub-events.}. We require models to generate the sub-event sequence for a previously unobserved process given observed processes along with their sub-event sequences, which we refer 
to as ``the observed process graphs" in the rest of this paper.
This task is more challenging than ``conventional" event predictions tasks, since it requires the {\em generation} of a sub-event sequence given a new, previously unobserved, process definition.

To address 
this problem, we propose an Analogous Process Structure Induction (APSI) framework.
Given a new process definition (e.g., `buy a house'), we first decompose it into two dimensions: predicate and argument. 
For each of these, we collect a group of processes that share the same predicate (i.e., `buy-\textit{ARG}') or same argument (i.e., `\textit{PRE}-house'), and then 
induce an abstract and probabilistic sub-event 
representation for each group.
Our underlying assumption 
is that processes that share
the same predicate or argument could be analogous to each other, and thus 
could share similar sub-event structures.
Finally, 
we merge these two abstract representations, using 
an instantiation module, to predict the sub-event structure of the target process.
By doing so, we only need a small number of analogous processes (as we show, 20, on average)
to generate 
unseen sub-events for the target process.
Intrinsic and extrinsic evaluations 
show that APSI outperforms all baseline methods and can generate meaningful sub-event sequences for unseen processes, which are proven to be helpful for predicting missing events.

The rest of the paper is organized as follows. 
Section~\ref{sec:APSI} introduces the Analogous Process structure induction (APSI) framework. Section~\ref{sec:evaluation} describes our intrinsic and extrinsic evaluation, demonstrating the effectiveness of APSI and the quality of the induced process knowledge. We discuss related works in Section~\ref{sec:related_work} and conclude this paper with Section~\ref{sec:conclusion}.

\section{The APSI Framework}\label{sec:APSI}

\begin{figure}
    \centering
    \includegraphics[width=\linewidth]{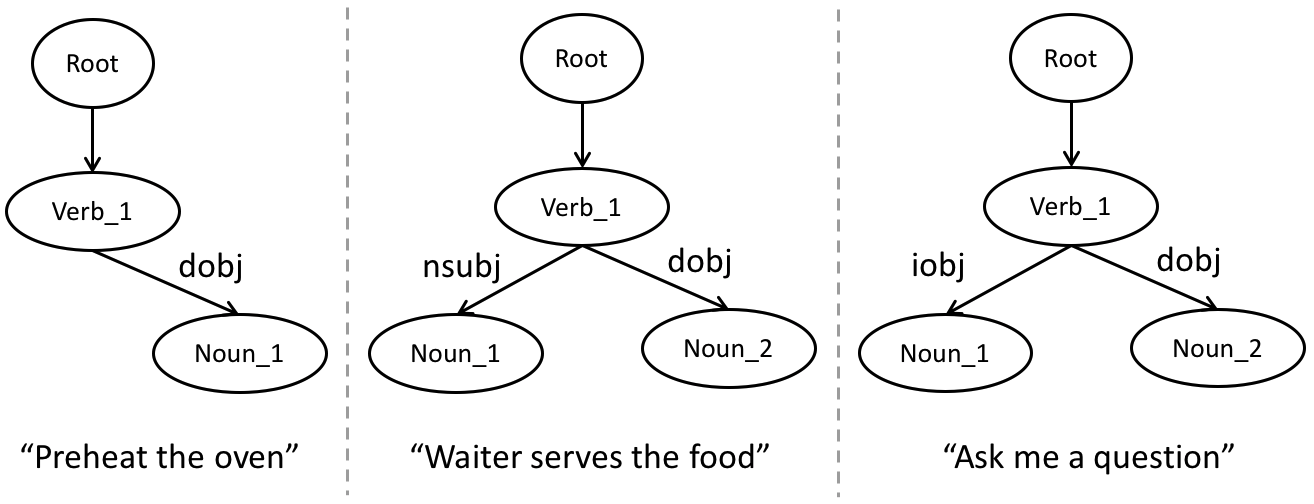}

    \caption{Examples of Sub-Event Representations.}
    \label{fig:event_example}
\end{figure}

Figure~\ref{fig:framework} illustrates the details of the proposed APSI framework. 
Given an unseen process $P$, a target sub-event sequence 
length $k$, and a set of observed process graphs $\GM$, the task is to predict 
a $k$-step sub-event sequence [$E^\prime_1, E^\prime_2, ..., E^\prime_k$] for $P$.
Each process graph $G \in \GM$ in the input contains a process definition $P^G$ and an $n$-step 
temporally ordered sub-event sequence [$E^G_1, E^G_2, ..., E^G_n$].
We assume that each process $P$ is described as a combination of a predicate and an argument (e.g., `buy+house') and each sub-event $E \in \EM$ is given as verb-centric dependency graph as used in~\cite{ASER2020} (see examples in  Figure~\ref{fig:event_example}).
In APSI, we 
decompose the target process into two dimensions (i.e., predicate and argument). 
For each target process, we collect a group of observed process graphs that share either the predicate or the argument with the target process; we assume that processes in these groups have sufficient information for predicting the structure of the target process.
We then leverage an event conceptualization module to induce an abstract representation of each process group. Finally, we merge the two abstract, probabilistic representations and instantiate it to generate a ground sub-event sequence as the final prediction.
Detailed descriptions of APSI components are introduced as follows.

\subsection{Semantic Decomposition}
Each process definition $P$ is given as a predicate and its argument, which we term below the two ``dimensions'' of the process definition.
We then collect all process graphs in $\GM$ that have the same predicate as $P$ into $\GM_p$ and those that have the same argument into $\GM_a$. We assume that these two sets provide the information needed to generate an abstract process representation that would guide the instantiation of the event steps for $P$.

\subsection{Semantic Abstraction}
The goal of the semantic abstraction step is to acquire abstract representations $S_p$ and $S_a$ for $\GM_p$ and $\GM_a$ respectively, to help transfer the knowledge from the grounded observed processes to the target new process.
To do so, we first need to conceptualize observed sub-events in $\GM_p$ and $\GM_a$ (e.g., ``eat an apple'') to a more abstract level (e.g., ``eat fruit'').
Clearly, each event could be conceptualized to multiple abstract events. 
For example, ``eat an apple'' can be conceptualized to ``eat fruit'' but also to ``eat food'', and the challenge is to determine the appropriate level of abstraction. On one hand, the conceptualized event cannot be too general, as we do not want to lose touch with the original event, and, on the other hand, 
if it is too specific, we will not aggregate enough instances of sub-events into it, thus we will have difficulties transferring knowledge to the new unseen process.
To automatically achieve the balance between these 
conflicting requirements and select the best abstract event for each observed sub-event, we model it as a weighted mutually exclusive set cover problem~\cite{DBLP:journals/corr/LuL14} and 
propose an efficient algorithm, described below, to solve it.
We then merge the repeated conceptualized events and determine their relative positions.

\subsubsection{Modeling Event Conceptualization}
For each event $E$, we first identify all potential events that it can be conceptualized to. 
If two sub-events $E_1$ and $E_2$ can be conceptualized to the same event $C$, we place $E_1$ and $E_2$ into the set $\EM_{C}$.
To qualitatively guide the abstraction process we introduce below a notion of {\em semantic loss} that we incur as we move up to more abstract representations. To measure the semantic loss during the conceptualization, we assign weight to each set:
\begin{equation}\label{eq:plausibility_score}
    W(\EM_{C}) = \frac{1}{\sum_{E \in \EM_{C}} F(E, C)},
\end{equation}
where $F(E, C)$ is a scoring function, defined below in Eq.~\ref{eq:conceptualization_score}, that captures the amount of ``semantic details" preserved due to abstracting from $E$ to $C$. 
With this definition, the event conceptualization problem can be formalized as finding exclusive\footnote{No sub-event can appear in two selected sets.} sets (such as $C$) that cover all observed events with minimum total weight.
In the rest of this section, we first introduce how to collect potential conceptualized events for each $E$, how we define $F$, and how we solve this discrete optimization problem. 

\noindent \textbf{Identifying Potential Conceptualizations}
Assume that sub-event $E$ contains $m$ words $w_1^E, w_2^E, ..., w_m^E$, each corresponds to a node in Figure~\ref{fig:event_example}; for each of these, we can retrieve a list of hypernym paths from WordNet~\cite{miller1998wordnet}.
For example, given the word ``house'', WordNet returns two hypernym paths\footnote{We omit the synset number for clear representation.}: (1) ``house''$\rightarrow$``building''$\rightarrow$``structure''$\rightarrow$...; (2) ``house''$\rightarrow$``firm''$\rightarrow$``business''$\rightarrow$....
As a result, we can find $\prod_{w \in E} L(w)$ potential conceptualized events for $E$, where $L(w)$ is the number of $w$'s hypernyms.
We denote the potential conceptualized event set for $E$ as $\CM_E$ and the overall set as $\CM$. 

\noindent \textbf{Conceptualization Scoring}
As mentioned above, for 
each pair of a sub-event $E$ and its potential conceptualization $C$,
we propose a scoring function $F(E, C)$ to measure how much 
``semantic information" is preserved after the conceptualization.
Motivated by~\citet{DBLP:journals/coling/BudanitskyH06} and based on the assumption that the more abstract
the conceptualized event is, 
the more semantic details are lost, we define $F(E,C)$ to be:
\begin{equation}\label{eq:conceptualization_score}
    F(E, C) = \prod_{i=1}^m w^{D(w_i^E, w_i^C)},
\end{equation}
\noindent
where $D(w_i^E, w_i^C)$ is the depth from $w_i^E$ to $w_i^C$ on the taxonomy path, and $w$ is a hyper-parameter\footnote{In practice, we use two separate hyper-parameters $w_v$ and $w_n$ for verbs and nouns, respectively.} measuring how much ``semantics" is preserved following each step of the conceptualization.

\begin{algorithm}[t]\caption{Event Conceptualization\label{alg:conceptualization}}
\textbf{INPUT:} Set of events $\EM$. Each $\E \in \EM$ is associated with a set of potential conceptualization events $\CM_E$. The overall conceptualized event set $\CM$. 
\begin{algorithmic}[1]
\small
\State Initialize event partition set $\mathcal{P} :=\emptyset$.
\While{$\EM \neq \emptyset$}
    \For{Each $ E \in \EM$}
        \For{Each $C \in \CM_E$}
            \State $\EM_C := \emptyset$.
            \State Compute $F(E, C)$ using Eq.~(\ref{eq:conceptualization_score}).
        \EndFor
    \EndFor
    \For{Each $ C \in \CM$}
        \For{Each $E \in \EM$}
            \If{$C \in \CM_E$}
                \State $\EM_C := \EM_C \cup \{E\}$.
            \EndIf
        \EndFor
        \State Compute $W(\EM_C)$ using Eq.~(\ref{eq:plausibility_score}).
    \EndFor
    \State Select $\EM_{C_{min}}$ with the minimum $W$ score.
    \State $\mathcal{\EM} := \EM \setminus \EM_{C_{min}}$
    \State $\mathcal{P} := \mathcal{P}\cup \{\EM_{C_{min}}\}$.
\EndWhile
\end{algorithmic}
\textbf{OUTPUT:} Partition of $n$ event subsets $\mathcal{P}=\{\EM_1, \EM_2, ..., \EM_n\}$, where each subset $\EM_i$ 
corresponds to a unique conceptualized event $C_i$.
\end{algorithm}


\noindent \textbf{Conceptualization Assignment}
Now we are able to model the procedure of finding proper conceptualized events as a weighted mutually exclusive set cover problem. Note that this is an NP-complete problem and requires a prohibitive computational cost to obtain the optimum solution~\cite{DBLP:conf/coco/Karp72}.
To obtain an efficient solution 
that is empirically sufficient for assigning conceptualized events with reasonable amount of instances, we develop a greedy procedure as described in Algorithm \ref{alg:conceptualization}.
For each retrieved process graph set $\GM_p$ or $\GM_a$, we collect all its sub-events as $\EM$ and use it as the input for the conceptualization algorithm.
In each iteration, we first compute the conceptualization score $F$ for all the ($E$, $C$) pairs and then compute the weight score for all conceptualization sets $\EM_C$.
After selecting the set with minimum weight,  $\EM_{C_{min}}$, we remove all the events covered by it 
from $\EM$ and repeat the process until no event is left.
After the conceptualization, we merge sub-events that are conceptualized to the same event and represent them with the resulting conceptualized event $C$, whose weight is defined to be $\overline{W}(C) = \frac{1}{W(\EM_C)}$.
Compared with the naive algorithm, which first expands all possible subsets (i.e., it includes all subsets of $\EM_C$ for all $C$) and then leverages the sort and filter technique to select the final subsets, we reduce the time complexity from $O(|\CM|\cdot|\EM|^2)$ to $O(n\cdot|\CM|\cdot|\EM|)$, where $n$ is the number of conceptualized events and is typically much smaller than $|\EM|$.

\subsubsection{Conceptualized Event Ordering}

After conceptualizing and merging all sub-events, we need to determine their loosely temporal order (e.g., whether they typically appear at the beginning or the end of these sub-event sequences). 
Let the set of selected conceptualized events 
be $\CM^*$. For each $C \in \CM^*$, we define its order score $T(C)$, 
indicating how likely $C$ is to appear first, as:
\begin{equation}
\small
    T(C) = \sum_{C^\prime \in \C^*} \theta (\sum_{E_C \in \EM_C} \sum_{E_{C^\prime} \in \EM_{C^\prime}} t(E_C, E_{C^\prime})-t(E_{C^\prime}, E_{C})),
\end{equation}
where $\theta$ is the unit step function and $t(E_C, E_{C^\prime})$ represents how many times $E_C$ appears before $E_{C^\prime}$ in an observed process graph.

\subsection{Sub-event Sequence Prediction}


In the last step, we leverage the two abstract representations we got for the 
predicate and argument of the target process definition to predict its final sub-events.
To do so, we propose the following instantiation 
procedure.
We are given the abstract representations $S_p$ and $S_a$, for the predicate and argument, respectively. Each is a set of conceptualized events associated with weights and order 
scores.
For each conceptualized event $C_p \in S_p$, using each event $C_a \in S_a$, we can generate a new instantiated event $\hat{C_p}$. For example, if $C_p$ is ``cut fruit'' and $C_a$ is `buy an apple', then our model 
would create the new event ``cut an apple''. 
Specifically, for each $w \in C_p$, if we can find a word $\hat{w}$ such that $\hat{w}$ is a hyponym of $w$, we will replace $w$ with $\hat{w}$ and repeat this process until no hyponym can be detected in $C_p$.
We denote the generated event by $\hat{C_p}$.
To account for the semantic loss during the instantiation procedure, we define the weight and order 
score of $\hat{C_p}$ as follows:
\begin{equation}\label{eq:new_weight}
\small
    \hat{W}(\hat{C_p}) = \overline{W}(C_p) \cdot F(\hat{C_p}, C_p) \cdot \frac{\sum_{C_a^\prime \in S_a} \overline{W}(C_a^\prime)}{\overline{W}(C_a)}
\end{equation}
\begin{equation}\label{eq:new_temporal}
\small
    \hat{T}(\hat{E_p}) = T(C_p) \cdot F(\hat{C_p}, C_p) \cdot \frac{\sum_{C_a^\prime \in S_a} \overline{W}(C_a^\prime)}{\overline{W}(C_a)},
\end{equation}
Similarly, we apply the same procedure to $C_a$ with $C_p$, and denote the resulted event  $\hat{C_a}$. 
We then repeatedly merge instantiated events by summing up their weights and averaging their order 
scores.
In the end, we select top $k$ sub-events based on the weights and sort them based on the order 
score as the sub-event sequence prediction.

\section{Evaluation}\label{sec:evaluation}

In this section, we conduct intrinsic and extrinsic evaluations to show that APSI can generate meaningful sub-event sequences for unseen processes, which can help predict the missing events. 
\subsection{Dataset}

We collect process graphs from the WikiHow website\footnote{https://www.wikihow.com.}~\cite{DBLP:journals/corr/abs-1810-09305}.
In WikiHow, each process is associated with a sequence of 
temporally ordered human-created steps.
For each step, as shown in Figure~\ref{fig:event_example}, we use the tool released by ASER~\cite{ASER2020} to extract events and construct the process graphs. 
We select all processes, where each step has one and only one event, and randomly split them into the train and test data. 
As a result, we got 13,501 training process graphs and 1,316 test process graphs\footnote{We do not need a development set because the proposed solution APSI is not a learning-based method.}, whose average sub-event sequence length is 3.56.

\begin{table*}[t]
\small
	\centering
    \subtable[Basic Setting (for each sub-event, we only predict and evaluate the verb)]{
      \begin{tabular}{l||cc|cc}
            \toprule 
            \multirow{2}{*}{Model} & \multicolumn{2}{c|}{String Match} & \multicolumn{2}{c}{Hypernym Allowed} \\
                & E-ROUGE1 & E-ROUGE2 & E-ROUGE1 & E-ROUGE2 \\
            \midrule
            Random & 2.9165& 0.4664 & 23.5873 & 8.1089 \\
            \midrule
            Seq2seq (GloVe) & 5.0323 & 1.4965 & 27.8710 & 13.0946\\
            Seq2seq (RoBERTa) & 4.5455 & 0.4831 & 28.0032 & 12.8502\\
            \midrule
            Top one similar process (Jaccard) & 8.8589 & 5.1000 & 28.6548 & 14.6231 \\
            Top one similar process (GloVe)  & 9.8797 & 5.1452 & 29.4203 & 13.6001 \\
            Top one similar process (RoBERTa) & 9.2599 & 4.7390 & 30.6599 & 15.8417 \\
            \midrule
            Analogous Process Structure Induction (APSI) & \textbf{14.8013} & \textbf{6.6045} & \textbf{36.1648} & \textbf{19.2418} \\
            \midrule
            Human & 29.0189 & 15.2542 & 50.4647 & 29.4423 \\
            \bottomrule
            \end{tabular}
    }
    \subtable[Advanced Setting (for each sub-event, we predict and evaluate all words)]{
      \begin{tabular}{l||cc|cc}
            \toprule 
            \multirow{2}{*}{Model} & \multicolumn{2}{c|}{String Match} & \multicolumn{2}{c}{Hypernym Allowed} \\
                & E-ROUGE1 & E-ROUGE2 & E-ROUGE1 & E-ROUGE2 \\
            \midrule
            Random & 0.0000 & 0.0000 & 0.5104 & 0.0903 \\
            \midrule
            Seq2seq (GloVe) & 0.1935 & 0.0534 & 0.9677 & 0.1069 \\
            Seq2seq (RoBERTa) & 0.4870 & 0.0000 & 1.7857 & 0.2899\\
            \midrule
            Top one similar process (Jaccard) & 0.6562 & 0.2257 & 2.4797 & 0.5867 \\
            Top one similar process (GloVe)  & 0.8750 & 0.2106 & 2.8801 & 0.7372 \\
            Top one similar process (RoBERTa) & 0.9479 & 0.3009 & 3.2811 & 0.9929 \\
            \midrule
            Analogous Process Structure Induction (APSI) & \textbf{3.4988} & \textbf{0.4513} & \textbf{6.1611} & \textbf{1.1885} \\
            \midrule
            Human & 11.6351 & 5.5905 & 18.0034 & 8.2695\\
            \bottomrule
            \end{tabular}
    }
	\caption{Intrinsic evaluation results of the induced process structures. On average, we have 1.7 human-generated sub-event sequences as the references for each test process. Best performing models are marked with the bold font. } \label{tab:main_result}
	\vspace{-0.1in}
\end{table*}

\subsection{Baseline Methods}

We compare with the following baseline methods:

\noindent \textbf{Sequence to sequence (Seq2seq):} One intuitive solution to the sub-event sequence prediction task would be modeling it as a sequence to sequence problem, where the process is treated as the input and the sub-event sequence the output. 
Here we adopt the standard GRU-based encoder-decoder framework~\cite{DBLP:conf/nips/SutskeverVL14} as the base framework and change the generation unit from words to events. 
For each process or sub-event, we leverage pre-trained word embeddings (i.e., GloVe-6b-300d~\cite{DBLP:conf/emnlp/PenningtonSM14}) or language models (i.e., RoBERTa-base~\cite{DBLP:journals/corr/abs-1907-11692}) as the representation, which are denoted as Seq2seq (GloVe) and Seq2seq (RoBERTa).

\noindent \textbf{Top One Similar Process:}
Another baseline is the ``top one similar process''. 
For each new process, we can always find the most similar observed process. 
Then we can use the sub-event sequence of the observed process as the prediction.
We employ different methods (i.e., token-level Jaccard coefficient or cosine similarity of GloVe/RoBERTa process representations) to measure the process similarity.
We denote them as Top one similar process (Jaccard), (GloVe), and (RoBERTa), respectively.

For each process, we also present a randomly generated sequence and a human-generated sequence\footnote{The human-generated sequence is randomly selected from the WikiHow and excluded during the evaluation.} as the lower-bound and upper-bound for sub-event sequence prediction models. 

\begin{figure*}[tb]
    \centering
	\includegraphics[width=0.95\linewidth]{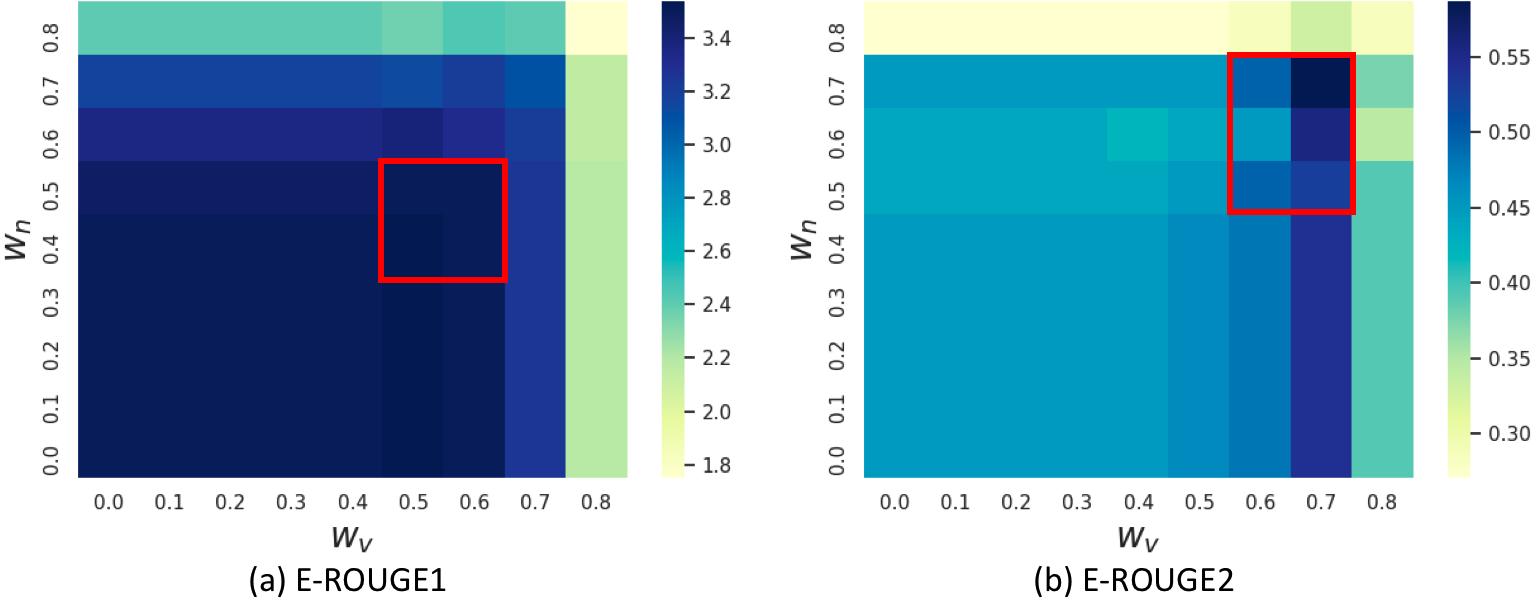}	
	\caption{Hyper-parameter influence on the quality of APSI generated sub-event sequences. For both $w_v$ and $w_n$, 0 indicates no conceptualization and the larger the value, the deeper the conceptualization is. Best performing ranges are marked with red boxes, which indicate that the suitable conceptualization level is the key to APSI's success. } 
	\label{fig:hyperparameter}
\end{figure*}

\subsection{Intrinsic Evaluation}

We first present the intrinsic evaluation to show the quality of the predicted sub-event sequences of unseen processes. For each test process, we provide the process name and the sub-event sequence length\footnote{We select the majority length of all references.} to evaluated systems and ask them to generate a fixed-length sub-event sequence.

\subsubsection{Evaluation Metric}


Motivated by the ROUGE score~\cite{lin-2004-rouge}, we propose an event-based ROUGE (E-ROUGE) to evaluate the quality of the predicted sub-event sequence.
Specifically, similar to ROUGE, which evaluates the generation quality based on N-gram token occurrence, we evaluate how much percentage of the sub-event and time-ordered sub-event pairs in the induced sequence is covered by the human-provided references.
We denote the evaluation over single event and event pairs as E-ROUGE1 and E-ROUGE2, respectively. 
We also provide two covering standards to better understand the prediction quality: (1) ``String Match'': all words in the predicted event/pairs must be the same as the referent event/pairs; (2) ``Hypernym Allowed'': the predicted and referent event must have the same dependency structure, and for the words on the same graph position, they should be the hypernym of or same as each other. For example, if the referent event is ``eat apple'' and the predicted event is ``eat fruit'', we still count it as a match.
The ``String Match'' setting is stricter, but the ``Hypernym Allowed'' setting also has its unique value to help better understand if our system is predicting relevant sub-events.

\subsubsection{Implementation Details}

In terms of 
training, we set both $w_v$ and $w_n$ to be 0.5 for our model. 
For the seq2seq baselines, we set the learning rate to be 0.001 and train the models until they converge on the training data.
All other hyper-parameters 
following the original paper.
In terms of the evaluation, we also provide two settings. (1) Basic: we follow previous works~\cite{DBLP:conf/lrec/GlavasSMK14} to predict and evaluate events based on verbs; (2) Advanced: we predict and evaluate events based on all words.

\subsubsection{Result Analysis}

We show the results in Table~\ref{tab:main_result}. In general, there is still a 
notable gap between current models' performance and human performance, 
but the proposed APSI framework can indeed generate 
sufficiently relevant sub-events. For example, if we only consider the verb. Even in the string match setting, 14.8\% of the predicted event and 6.6\% of the ordered event pairs are covered by the references, which is much better than the random guess and nearly half of the performance of human beings. If hypernym is allowed, 36\% and 19\% of the predicted event and event pairs are covered. 
Besides that, if we take all words in the event into consideration, the task becomes more challenging. 
Specifically, even human 
can only achieve 11.63 E-ROUGE1 and 5.59 E-ROUGE2, which suggests that low scores achieved by current models are probably due to the limitation of the current dataset (e.g., on average, we only have 1.7 references for each test process). 
If more references are provided, the performance of all models will also increase.
In the rest of the intrinsic evaluation, we present more detailed analysis based on the advanced setting (string match) and a case study to help better understand the performance of APSI.

\subsubsection{Effect of the Instantiation Module}

One key step in our framework is how to leverage the two abstract representations to predict the final sub-event sequence.
In APSI, we propose an instantiation module, which jointly leverages the two representations to generate detailed events. To show its effect, we compare it with two other options: (1) Simple Merge: Merge two representation and select the top $k$ sub-events based on the weight; (2) Normalized: First normalize the weight of all sub-events based on each representation and then select the top $k$ sub-events.

\begin{table}[t]
\small
    \centering
    \begin{tabular}{l||c|c}
    \toprule
        Model & E-ROUGE1  & E-ROUGE2  \\
        \midrule
        Simple Merge & 2.5884 & 0.4062 \\
        Normalized & 2.2238 & 0.3611 \\
         \midrule
        APSI (Instantiation) & \textbf{3.4988} & \textbf{0.4513} \\
         \bottomrule
    \end{tabular}
    \caption{Performance of different merging methods.}
    \label{tab:merging}
\end{table}

From the result in Table~\ref{tab:merging}, we can see that due to the imbalanced distribution of the two representations, simply choosing the most weighted sub-events is problematic. 
On average, for each predicate, we can collect 18.04 processes, while we can only collect 1.92 processes for each argument. As a result, the sub-events in the predicate representation typically have a larger weight. Thus if we simply merge them, most of the predicted sub-events will come from the predicate representation.
Ideally, the ``normalized'' method can eliminate the influence of such imbalance, but it also amplifies the noise and achieves worse empirical performance.
Differently, the proposed instantiation module uses events in one representation as the reference to help instantiate the events in the other one. As a result, we jointly use these two representations to generate a group of detailed events, and then we can select the top $k$ generated new events. 
By doing so, we do not only go detailed from the abstract representation but also avoid the imbalanced distribution issue.



\begin{figure}
	\centering
	\includegraphics[width=\linewidth]{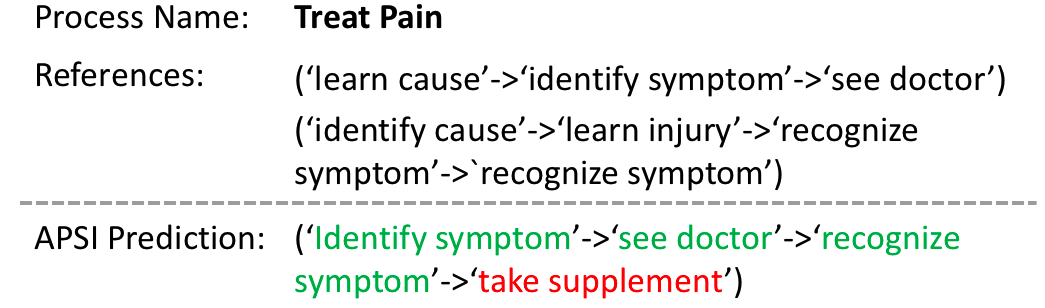}
	\caption{Case Study. We mark the covered and not covered predictions with green and red colors.}
	\label{fig:case_study}
\end{figure}

\subsubsection{Hyper-parameter Analysis}

In APSI, we use two hyper-parameters $w_v$ and $w_n$ to control the conceptualization and instantiation depth we want over verbs and nouns respectively.
0 means no conceptualization and the larger value indicates more conceptualization we encourage.
We show the performance of APSI with different hyper-parameter combinations in Figure~\ref{fig:hyperparameter}, from which we can see that a suitable level of conceptualization is the key to the success of APSI.
If no conceptualization is allowed, all the predicted events are restricted to the observed sub-event, thus we cannot predict ``search house'' after seeing ``search car'' and some events about the house.
On the other hand, if we do not restrict the depth of conceptualization, all the sub-events will be conceptualized to be too general. 
As a result, even with the instantiation module, we could not predict the detailed sub-event as we want.

\subsubsection{Case Study}
Figure~\ref{fig:case_study} shows an example that we use to analyze the current limitations of APSI. 
We can see that APSI can successfully predict events like ``identify symptoms'', but fails to predict event ``identify causes''. Instead, it predicts ``take supplements''. This is because APSI learns to predict such sequence from other processes like ``treat diarrhea'' or other diseases in the observed process graphs. Treating those 
diseases typically does not involve identifying the cause, which is not the case for treating pain. 
And, treating diseases often involves taking medicines, which can be conceptualized to ``take supplement''. As no events about pain helps instantiate ``supplement", APSI just predicts it. 

\subsection{Extrinsic Evaluation}

As discussed by~\cite{rumelhart1975notes}, the knowledge about process and sub-events can help understand event sequences.
Thus, in this section, we investigate whether the induced process knowledge can help predict the missing events.
Given a sub-event sequence, for each event in the sequence, we can use the rest of the sequence as the context and ask models to select the correct event against one negative event example.
To make the task challenging, instead of random sampling, we follow 
\citet{DBLP:conf/acl/ZellersHBFC19} to select similar but wrong negative candidates based on their representation (i.e., BERT~\cite{DBLP:conf/naacl/DevlinCLT19}) similarity.
We use the same training and test as the intrinsic experiment and as a result, we got 13,501 training sequences and 7,148 test questions.

\begin{figure}
    \centering
    \includegraphics[width=\linewidth]{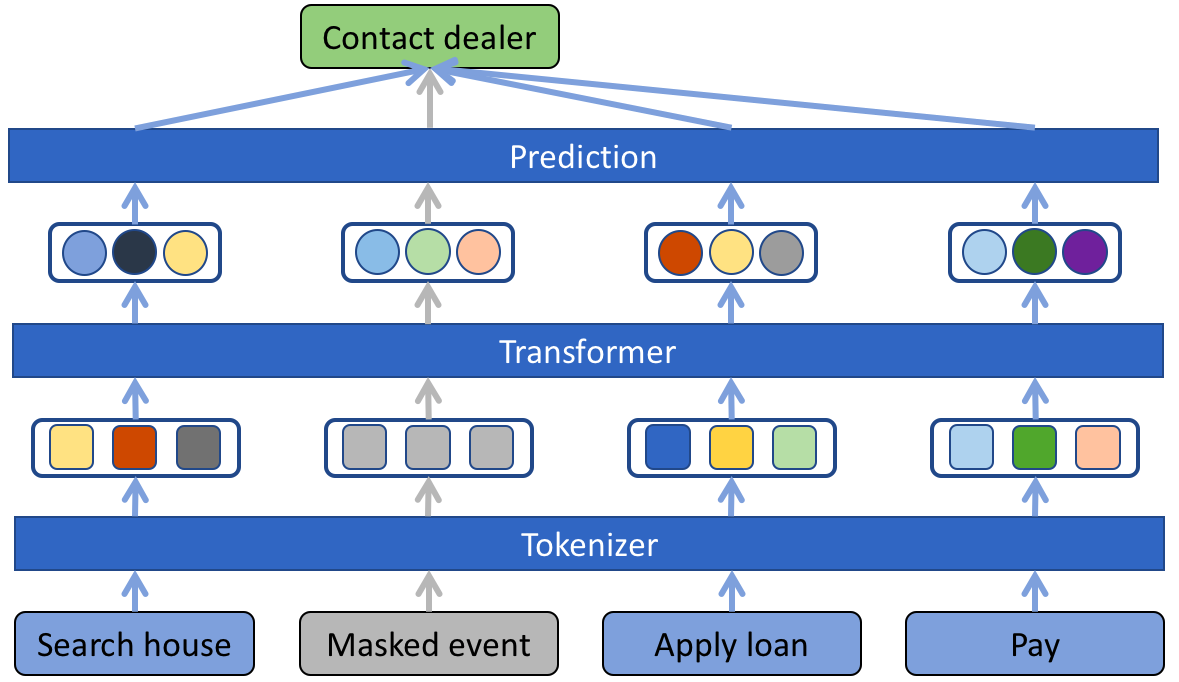}
    \caption{Demonstration of the event masked LM. Pre-trained language models are trained to predict the masked event given other events as the context.}
    \label{fig:event_LM}
\end{figure}

The baseline method we are comparing with is the event-based masked language model\footnote{On our dataset, the RoBERTa based event LM model outperforms existing LSTM-based event prediction models.}, 
whose demonstration is shown in figure~\ref{fig:event_LM}. 
We use pre-trained RoBERTa-base~\cite{DBLP:journals/corr/abs-1907-11692} to initialize the tokenizer and transformer layer and all sequences of training processes as the training data.
To show the value of understanding the relationship between process and their sub-event sequence, for each sub-event sequence in the test data, we first leverage the process name and different structure prediction methods to predict sub-event sequences and use them as additional context to help the event masked LM to predict the missing event.
To show the effect upper bound of adding process knowledge, we also tried adding the process structure provided by human beings as the context\footnote{We randomly select another sub-event sequence that describes the same process from WikiHow, which could be different from the currently tested sequence. As a result, adding such sequence cannot help predict all missing events.}, which is denoted as `+Human'. All models are evaluated based on accuracy.

\begin{table}[t]
\small
    \centering
    \begin{tabular}{l||c|c}
    \toprule
        Model & Accuracy  & $\Delta$  \\
        \midrule
        RoBERTa-based Event LM & 73.59\% & - \\
        \midrule
        \quad + Seq2seq (GloVe) & 73.06\% & -0.53\% \\
        \quad + Seq2seq (RoBERTa) & 72.33\% & -1.26\% \\
        \midrule
        \quad + Top1 similar (Jaccard) & 72.76\% & -0.83\% \\
        \quad + Top1 similar (GloVe) & 74.14\% & 0.55\% \\
        \quad + Top1 similar (RoBERTa) & 74.16\% & 0.57\% \\
         \midrule
        \quad + APSI & 74.78\%$^\dagger$ & 1.19\% \\
        \midrule
        \quad + Human & 76.97\%$^\ddagger$ & 3.38\% \\
         \bottomrule
    \end{tabular}
    \caption{Results on the event prediction task. $\dagger$ and $\ddagger$ indicate the statistical significance over the baseline with p-value smaller than 0.01 and 0.001 respectively.}
    \label{tab:extrinsic}
\end{table}

From the results in Table~\ref{tab:extrinsic}, we can make the following observations. First, adding high-quality process knowledge (i.e., APSI and Human) can significantly help the baseline model, which indicates that adding knowledge about the process can help better understand the event sequence. Second, the effect of process knowledge is positively correlated with their quality as shown in Table~\ref{tab:main_result}. Adding a low-quality process structure may hurt the performance of the baseline model due to the introduction of the extra noise. Third, the current way of using process knowledge is still very simple and there is room for better usage of the process knowledge, as the research focus of this paper is predicting process structure rather than applying it, we leave that for the future work.

\section{Related Works}\label{sec:related_work}

Throughout history, considering the importance of events in understanding human language (e.g., commonsense knowledge~\cite{DBLP:conf/ijcai/ZhangKSR20}), many efforts have been devoted to define, represent, and understand events.
For example, VerbNet~\cite{schuler2005verbnet} created a verb lexicon to represent the semantic relations among verbs.
After that, FrameNet~\cite{newBakerFiLo98} proposed to represent the event semantics with schemas, which has one predicate and several arguments.
Apart from the structure of events, understanding events by predicting relations among them also becomes a popular research topic (e.g., TimeBank~\cite{pustejovsky2003timebank} for temporal relations and Event2Mind~\cite{DBLP:conf/acl/SmithCSRA18} for causal relations).
Different from these \textit{horizontal} relations between events, in this paper, we propose to understand event \textit{vertically} by treating each event as a process and trying to understand what is happening (i.e., sub-event) inside the target event.
Such knowledge is also referred to as event schemata~\cite{zacks2001event} and shown crucial for how humans understand events~\cite{abbott1985representation}.
One line of related works in the NLP community is extracting super-sub event relations from textual corpus~\cite{DBLP:conf/naacl/HovyMVAP13,DBLP:conf/lrec/GlavasSMK14}.
The difference between this work and them is that we are trying to understand events by directly generating the sub-event sequences rather than extracting such information from text.
Another line of related works is the narrative schema prediction~\cite{DBLP:conf/acl/ChambersJ08}, which also holds the assumption that event schemata can help understand events. But their research focus is using the overall process implicitly to help predict future events while this work tries to understand events by knowing the relation between processes and their sub-event sequences explicitly.



\section{Conclusion}\label{sec:conclusion}

In this paper, we try to understand events \textit{vertically} by viewing them as processes and predicting their sub-event sequences.
Our APSI framework 
is motivated by the notion of analogous processes, and attempts to transfer knowledge from (a very small number of) familiar processes to a new one. 
The intrinsic evaluation demonstrates the effectiveness of APSI and the quality of the predicted sub-event sequences.
Moreover, the extrinsic evaluation shows that, even with a naive application method, the process knowledge can help better predict missing events.

\section*{Acknowledgements}
This research is supported by the Office of the Director of National Intelligence (ODNI), Intelligence Advanced Research Projects Activity (IARPA), via IARPA Contract No. 2019-19051600006 under the BETTER Program, and by contract FA8750-19-2-1004 with the US Defense Advanced Research Projects Agency (DARPA). The views expressed are those of the authors and do not reflect the official policy or position of the Department of Defense or the U.S. Government.
This paper is also partially supported by  Early Career Scheme (ECS, No.
26206717), General Research Fund (GRF, No. 16211520), and Research
Impact Fund (RIF, No. R6020-19) from the Research Grants Council (RGC)
of Hong Kong.

\bibliography{ccg,cited,new}
\bibliographystyle{acl_natbib}

\clearpage


\end{document}